\documentclass{IEEEtran}
\usepackage{textcomp}
\usepackage[latin9]{inputenc}
\usepackage[active]{srcltx}
\usepackage{array}
\usepackage{float}
\usepackage{amsmath}
\usepackage{graphicx}
\usepackage{geometry}
\makeatletter

\providecommand{\tabularnewline}{\\}

\usepackage{amsfonts}\usepackage{algorithmic}
\usepackage{algorithm}
\usepackage{array}
\usepackage{stfloats}
\usepackage{url}
\usepackage{verbatim}
\usepackage{cite}
\makeatother

\begin{document}
\title{Volumetric Reconstruction From Partial Views for Task-Oriented Grasping}
\author{Fujian Yan, Hui Li, and Hongsheng He$^{*}$\thanks{Fujian Yan is with the School of Computing, Wichita State University,
Wichita, KS, 67260, USA}\thanks{Hui Li and Hongsheng He are with the Department of Computer Science,
The University of Alabama, Tuscaloosa, AL 35487 USA.}\thanks{$^{*}$Correspondence should be addressed to Hongsheng He, hongsheng.he@ua.edu.}}
\maketitle
\begin{abstract}
Object affordance and volumetric information are essential in devising
effective grasping strategies under task-specific constraints. This
paper presents an approach for inferring suitable grasping strategies
from limited partial views of an object. To achieve this, a recurrent
generative adversarial network (R-GAN) was proposed by incorporating
a recurrent generator with long short-term memory (LSTM) units for
it to process a variable number of depth scans. To determine object
affordances, the AffordPose knowledge dataset is utilized as prior
knowledge. Affordance retrieving is defined by the volume similarity
measured via Chamfer Distance and action similarities. A Proximal
Policy Optimization (PPO) reinforcement learning model is further
implemented to refine the retrieved grasp strategies for task-oriented
grasping. The retrieved grasp strategies were evaluated on a dual-arm
mobile manipulation robot with an overall grasping accuracy of 89\%
for four tasks: lift, handle grasp, wrap grasp, and press.
\end{abstract}

\begin{keywords} 3D volumetric composition, task-oriented grasping,
generative adversary network, and views adaptive model. \thispagestyle{empty}
\end{keywords}

\section{Introduction}

Traditional robotic manipulation is  typically designed for repetitive
tasks involving identical objects. However, with the increasing deployment
of robots in human-centric environments, there is a growing need for
these systems to exhibit greater flexibility and adaptability, particularly
in grasping strategies for previously unseen objects. For instance,
assistive robots assembling furniture, such as chairs, must adjust
their grasp based on the unique geometry of each component. Humans
naturally estimate an object's volume to determine optimal grasp points
to ensure a stable and secure hold. Emulating this capability in robots
can enhance their dexterity and effectiveness in handling diverse
objects. This paper aims to replicate human-like volume assessment
in robotic manipulation to enable stable grasping of various objects
with partial views.

To develop a reliable grasping strategy, physics-based analysis methods
have been proposed \cite{rodriguez2012caging,weisz2012pose}. These
approaches typically assume complete knowledge of object properties,
an assumption that does not hold, where objects are often unfamiliar
or previously unseen \cite{goldfeder2011data}. To obtain the physic
attributes of objects like volume, a grasping plan based on synthetic
point clouds was introduced in \cite{mahler2017dex}. However, this
method relies on 2.5D point clouds with comprehensive sensor coverage,
making it unsuitable for scenarios where only partial object views
are available. Shape-completion-based approaches have been explored
\cite{varley2017shape}, but these methods do not account for task-specific
grasping constraints. Integrating object affordances into grasp planning
can enhance task-aware grasping strategies, improving adaptability
in real-world applications.

The problem of 3D volumetric reconstruction is essential for deriving
3D models from 2D or 2.5D sensing data. It aims to reconstruct the
3D surfaces and fill in the unobserved volumes beneath those surfaces
\cite{runz2020frodo}. Unlike 3D reconstruction or 3D shape generation,
which primarily focus on surface reconstruction, 3D volumetric reconstruction
seeks to create complete models encompassing volumetric, shape, and
dimensional information. Such comprehensive models are valuable in
various applications, including robotic manipulation \cite{9562073},
object deformation \cite{yin2021modeling}, robotic scene understanding
\cite{9635939}, and human-robot collaboration \cite{green2008human}.
Traditional 3D reconstruction methods based on feature matching and
view registration, such as structure from motion (SfM) \cite{yi2014survey}
and visual simultaneous localization and mapping (SLAM) \cite{silveira2008efficient},
usually require a large number of scans from different perspectives
with dense features \cite{10.1145/2508363.2508374}. It is challenging
for these methods to reconstruct a 3D model with limited scans. Due
to the restricted point of view (POV) of an RGB-D sensor, it is often
impractical to cover all surfaces with just a few scans. Consequently,
the reconstructed 3D models may miss significant components. Hole-filling
techniques, such as Laplacian hole filling \cite{sorkine2004laplacian}
and Poisson surface reconstruction \cite{kazhdan2006poisson}, can
handle minor holes in the reconstructed models but face challenges
in processing or generating complex structures.
\begin{figure*}
\begin{centering}
\includegraphics[scale=0.9]{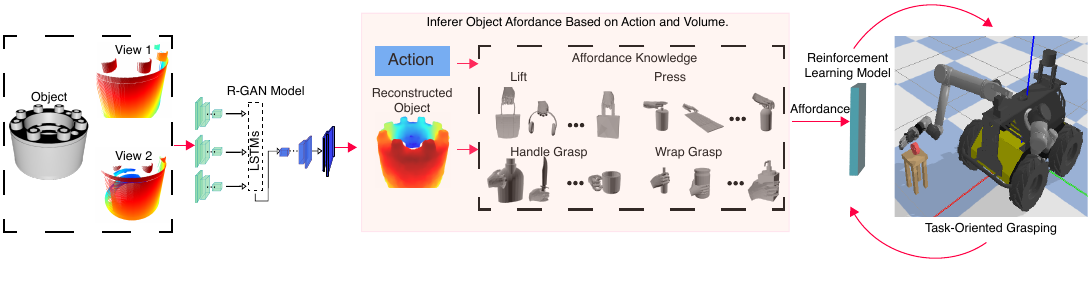}
\par\end{centering}
\begin{centering}
\caption{Task-oriented grasping based on volumetric reconstruction and affordance
knowledge.\label{fig:title_image} }
\par\end{centering}
\end{figure*}

In this paper, we proposed an approach for task-oriented grasping
of unseen objects using constructed object volumes and affordance
knowledge. To achieve this, we designed a recurrent generative adversarial
(R-GAN) model to reconstruct object volumes from limited scans, akin
to how humans reconstruct object volume from slight peaks. The R-GAN
model learns a smooth function that maps depth scans to volumetric
reconstructions. The model integrates a recurrent generator with a
3D convolutional encoder-decoder structure augmented by Long Short-Term
Memory (LSTM) units. The encoder extracts features that represent
stereoscopic structures from the depth scans and compresses them into
a latent space. Leveraging LSTM\textquoteright s memorization ability,
the model can process a single or a sequential depth scans. The decoder,
followed by LSTM units, decodes these features back into a 3D representation.
The discriminator, composed of 3D convolutional layers, compares the
reconstructed model with the object\textquoteright s point cloud.
To predict the optimal grasping strategy, we designed a grasping
approximation model that takes the volume of objects as input, then
computes the similarity against objects in AffordPose knowledge dataset,
which is a large-scale dataset of task-oriented hand-object interactions
with affordance-driven hand poses \cite{Jian_2023_ICCV}. This knowledge
base provides task-based grasp strategies informed by object affordances.
 The retrieved grasping strategy are refined by reinforcement learning
approach \cite{9562073}, and the proposed approach was evaluated
on a dual-arm robot. 

The major contributions of the paper include: 1) The design of the
R-GAN model for reconstructing volume of objects rather than just
the 3D shape. 2) Enabling robots to learn volumetric knowledge of
unseen objects.

\section{Volume Reconstruction for Task-Oriented Grasping\thispagestyle{empty}}

As illustrated in Fig.~\ref{fig:title_image}, the framework of
task-oriented grasping based on volumetric reconstruction and affordance
knowledge comprises three key components: volumetric reconstruction,
affordance-based grasp inference, and reinforcement learning. Limited
depth scans are collected to reconstruct the object's volume. The
action (e.g. lift, grasp) and the reconstructed volume are used to
find similar objects in the affordance knowledge base to infer object
affordance. As affordance decides the grasping pose for a task, it
may not fit the real-world object accurately. A reinforcement learning
model is therefore developed to refine the grasp for optimal grasping
strategies.

While RGB-D sensors are widely employed for capturing depth information,
a single RGB-D sensor provides depth data from only a single point
of view (POV), limiting the ability to reconstruct a complete 3D model
\cite{wang2014robust}. Although multiple RGB-D sensors can be utilized
to capture depth information from different perspectives, practical
constraints often make panoramic scanning infeasible in real-world
applications \cite{wang2015mobile}. 

We designed a Recurrent Generative Adversarial Network (R-GAN) to
generate a 3D volumetric model from a single or a limited sequence
of depth scans. The model processes these scans through an encoder
that extracts sequential features $S_{1},\,S_{2},\ldots,\,S_{n}$,
which are then fed into a LSTM layer for temporal feature learning.
The encoder structure is dynamically adapted based on the number of
available depth scans to ensure flexibility in different scanning
scenarios. The proposed R-GAN model allows flexible integration of
single or multiple depth scans, thereby enabling robust 3D object
reconstruction under varying conditions.

\subsection{3D Volumetric Reconstruction\thispagestyle{empty}}

The main challenge for 3D volumetric reconstruction is generating
unknown points beneath the surface. To solve this problem, we extends
the vanilla GAN model to compose unknown points. The recurrent generator
in the R-GAN model can take sequential depth inputs to reconstruct
3D volumetric models $\hat{Y}$. The discriminator takes both $\hat{Y}$
and ground truth $Y$ as inputs to differentiate them.

The designed model can adaptively take a single depth scan or a few
multiple depth scans $\{X_{\mathrm{i}}|X_{\mathrm{i}}\in\mathbb{R^{\mathrm{m\times m\times m}}},\mathrm{i}=1,2,3,...,n\}$.
The model maps input data $X_{i}$ to a latent space to approximate
the volume. To get accurate 3D volumetric reconstruction, the model
approximates parameters $\theta_{\mathrm{gen}}$ for the generator
and $\theta_{\mathrm{dis}}$for the discriminator by 
\begin{align}
\mathrm{\theta_{gen}^{*},\theta_{dis}^{*}=\underset{\theta_{gen}}{min}\underset{\theta_{dis}}{max}(\mathit{f}_{\mathrm{gen}}(\theta_{\mathrm{gen}},\thinspace\mathnormal{X}_{1},\dots,X_{i}),}\nonumber \\
f_{\mathrm{dis}}(\theta_{\mathrm{dis}},\thinspace Y_{j},f_{\mathrm{gen}}(\cdot)))\label{eq:loss_function}
\end{align}
where $Y_{\mathrm{j}}$ is the corresponding ground truth for a single
depth scan $X$ or multiple-views depth scans $X_{\mathrm{i}}$. $\theta_{\mathrm{gen}}^{*}$
is optimized parameters for the generator, and $\theta_{\mathrm{dis}}^{*}$
is the optimized parameter for the discriminator. The model contains
a generator $f_{\mathrm{gen}}(\cdot):X_{\mathrm{i}}\rightarrow\hat{Y}$
that reconstructs 3D volumetric model $\hat{Y}$ from the inputs,
and a discriminator $f_{\mathrm{dis}}(\cdot)$ to determine how different
between the $\hat{Y}$ and ground truth $Y$. Sequential depth scans
$\{X_{\mathrm{i}}|X_{\mathrm{i}}\in\mathbb{R}^{3},i=1,2,3,...,n\}$
are taken by the generator 
\begin{equation}
\mathrm{\mathit{f}_{gen}:\boldsymbol{\hat{\mathit{Y}}}=\sigma\sideset{(}{_{i=1}^{N}}\sum\boldsymbol{\mathit{W}}_{gen}^{i}\boldsymbol{\mathit{X}}_{i}+\boldsymbol{\mathit{b}}_{gen}^{i})}\label{eq:generator}
\end{equation}
where $\boldsymbol{W}_{\mathrm{gen}}^{\mathrm{i}}$ and $b_{\mathrm{gen}}^{\mathrm{i}}$
are weights and bias for each encoder in the generator network, respectively.
The $\mathrm{i}$ is the indicator for a specific number of the encoders
for adaptively taking the input scans. To extract features from an
input data, we designed an encoder that is constructed by a five-layer
3D convolutional neural network (3D-CNN).

The vanilla GAN model can take a single-depth scan of an object as
input to approximate 3D volumetric information. However, when provided
with multiple depth scans, it treats each scan as an independent input
To train the model with multiple depth scans, we extends the generator
with LSTM units. To generate sequential inputs for LSTM units, each
extracted feature tensor $x_{\mathrm{t}}$ from each encoder is flatten
to $\bar{x}_{\mathrm{t}}^{\mathrm{k}}$, where $\mathrm{k}$ is the
view index. The concatenated sequential tensor $\boldsymbol{X}_{\mathrm{t}}=\{\bar{x}_{\mathrm{t}}^{1},\bar{x}_{\mathrm{t}}^{2},\dots,\bar{x}_{\mathrm{t}}^{n}\}$,
which has a dimension of $(\mathrm{m\times t\times f})$. The $\mathrm{m}$
is number of depth scans, $\mathrm{t}$ is the time-step, and $\mathrm{f}$
is the number of elements in each feature. The LSTM units in the proposed
method consists of three gates governing data flow from the encoder
to the decoder in the generator. The input gate $i_{\mathrm{t}}$
control the data flow from the input to the hidden state $h_{\mathrm{t}}$,
the forget gate $f_{\mathrm{t}}$ controls the data flow from previous
hidden state $h_{\mathrm{t-1}}$ to current hidden state $h_{\mathrm{t}}$,
the operation of an LSTM unit can be described as 
\begin{equation}
\begin{array}{l}
\mathrm{\mathit{i}_{t}=\sigma(\mathit{\boldsymbol{W}}_{i}\cdot[\mathit{\boldsymbol{X}}_{t},\mathit{\boldsymbol{h}}_{t-1}]+\boldsymbol{\mathit{b}}_{i})}\\
\mathrm{\mathit{f}_{t}=\sigma(\mathit{\boldsymbol{W}}_{f}\cdot[\boldsymbol{\mathit{X}}_{t},\boldsymbol{\mathit{h}}_{t-1}]+\boldsymbol{\mathit{b}}_{f})}\\
\mathrm{\mathit{o}_{t}=\sigma(\boldsymbol{\mathit{W}}_{o}\cdot[\mathit{\boldsymbol{X}}_{t},\mathit{\boldsymbol{h}}_{t-1}]+\boldsymbol{\mathit{b}}_{o})}\\
\mathrm{\mathit{s}_{t}=\mathit{f}_{t}\odot\mathit{s}_{t-1}+\mathit{i}_{t}\odot\mathrm{tanh}(\boldsymbol{\mathit{W}}_{s}\cdot[\boldsymbol{\mathit{X}}_{t},\mathit{\boldsymbol{h}}_{t-1}]+\boldsymbol{\mathit{b}}_{s})}\\
\mathrm{\mathit{h}_{t}=\mathit{o}_{t}\odot\mathrm{tanh}(\mathit{s}_{t})}
\end{array}\label{eq:LSTM}
\end{equation}
where $\boldsymbol{W}_{i},\boldsymbol{W}_{f},\boldsymbol{W}_{o},\boldsymbol{W}_{s}$
are weights for the input gate, the forget gate, the output gate,
and the memory cell, respectively. The terms $b_{i},b_{f},b_{o},b_{s}$
are bias for the input gate, the forget gate, the output gate, and
the memory cell, respectively. The memory cell is represented by $s_{\mathrm{t}}$.
We used $\odot$ to represent element-wise multiplication.

The inputs $X_{i}$ are fed into the 3D convolutional encoder with
five 3D convolutional layers. In contrast to generators in vanilla
GAN models, which generate data randomly. The recurrent generator
in the designed model can generate data based on depth views. Two
fully connected layers are used to flat these extracted features from
the encoder to feed into the LSTM units. Also, these extracted features
are sequentially concatenated. LSTM units are followed by a five layers
3D-CNN decoder and a two layers 3D-CNN up-scaling network.

The discriminator can evaluate the difference between the reconstructed
3D volumetric model $\hat{Y}\in\mathbb{R^{\mathrm{m\times m\times m}}}$
and the 3D CAD model (ground truth) $Y\in\mathbb{R}^{\mathit{\mathrm{m\times m\times m}}}$
of an object. The difference between the average output of the discriminator
for the reconstruction model $\hat{Y}$ and the ground truth $Y$
is calculated by 
\begin{equation}
m=\sideset{\frac{1}{K}}{_{1}^{K}}\sum\mathrm{\mathit{f}_{dis}}(\text{\ensuremath{\hat{Y})-\sideset{\frac{1}{J}}{_{1}^{J}}\sum\mathrm{\mathit{f}_{dis}}(Y)}}\label{eq:mean_features}
\end{equation}
where $K$ and $J$ are the number of elements in $\mathrm{\mathit{f}_{dis}}(\hat{Y})$
and $\mathrm{\mathit{f}_{dis}(Y)}$ respectively. The term $\mathrm{\mathit{f_{\mathrm{dis}}}}(\cdot)$
is a 3D-CNN contains six 3D convolutional layers, and the first five
3D convolutional layers are activated by ReLU and the last layer takes
the sigmoid function as the activation function.

In order to train the model, we generate a substantial amount of training
and testing data with artificially rendered depth scans of an object
and its complete point cloud. Each object is placed in the Blender
environment, and the ambient light and geometrical structures of the
environment are the same. A synthetic RGB-D camera is placed in the
environment and takes sequential depth scans of the object in the
first and second quadrants in the x-y plane. The synthetic RGB-D camera
is placed 1.6 meters away from the object. The resolution of each
depth scan is 512$\mathnormal{\times}$512.

\subsection{Task-Oriented Grasping with Knowledge of Object Affordance}

Object affordance refers to the potential actions an object offers
based on its shape, material, and function. The function defines the
intended action (e.g., grasping, pouring), while the shape determines
geometric compatibility with the manipulator. An object affords an
action if its shape, size, and orientation align with task constraints.
For example, a cup affords pouring due to its open top (task relevance)
and graspable shape (volumetric match). In robotics, understanding
affordance enables robots to interact with objects effectively, facilitating
grasping and manipulation without explicit instructions.

Determining object affordance is challenging due to variations in
design, context dependence, and ambiguity. Objects within the same
category differ in shape, requiring generalization while ensuring
precise interaction. Affordances also depend on orientation and environment,
as a chair affords sitting only when upright. Ambiguous affordances,
such as a screwdriver\textquoteright s multiple uses, further complicate
decision-making.

\begin{figure}[h]
\begin{centering}
\includegraphics[width=0.9\columnwidth]{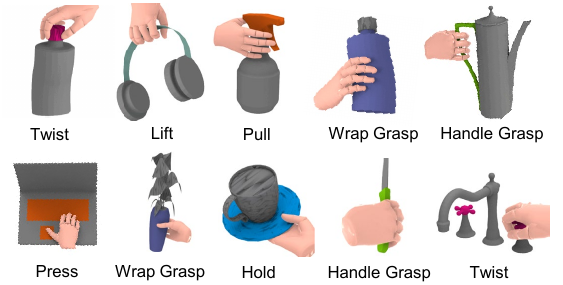}
\par\end{centering}
\caption{Part-level affordances are annotated, and the corresponding grasping
strategy is then matched to each affordance. \label{fig: partlevel}}
\end{figure}

In this paper, we infer object affordance from the AffordPose knowledge
dataset \cite{Jian_2023_ICCV}, which provides part-level affordance
labeling and a corresponding grasping strategy, as shown in Fig.~\ref{fig: partlevel}.
The annotated grasp strategy specifies the grasp pose, along with
the relative position and orientation between the hand and the object
part, enabling more adaptive and informed robotic grasping. 

To improve efficiency, the AffordPose dataset was reorganized into
task categories (e.g., handle grasp, wrap grasp, lift, press). The
framework first identified the task category $c$ and then computed
the volume similarity $d'$. The object with the smallest $d'$ was
selected for grasping strategy transfer
\begin{equation}
d'=\min_{O_{i}\in c}d_{\text{CD}}(O_{r},O_{i})
\end{equation}
where $O_{r}$ is the reconstructed object, and $O_{i}$ is the $i^{\text{th}}$
object under category $c$. The Chamfer Distance (CD), $d_{\text{CD}}$,
is defined as
\begin{equation}
d_{\text{CD}}(O_{r},O_{i})=\sum_{p\in O_{r}}\min_{q\in O_{i}}\|p-q\|^{2}+\sum_{q\in O_{i}}\min_{p\in O_{r}}\|q-p\|^{2}
\end{equation}
where $p$ is a point in the reconstructed object's point cloud $O_{r}$,
and $q$ is a point in the reference object's point cloud $O_{i}$.
The term $\min_{q\in O_{i}}\|p-q\|^{2}$ finds the closest point in
$O_{i}$ for each point in $O_{r}$, while $\min_{p\in O_{r}}\|q-p\|^{2}$
finds the closest point in $O_{r}$ for each point in $O_{i}$. The
summation aggregates these distances to quantify the overall shape
difference.

The retrieved grasping strategy may not fit the reconstructed object
well. To further adapt it to the reconstructed object and ensure stability,
a Proximal Policy Optimization (PPO) reinforcement learning model
from our previous work \cite{9562073} was employed. The PPO model
refines the retrieved grasping strategy by iteratively adjusting the
grasp location and grasp pose to better conform to the reconstructed
object's geometry and properties. For each grasping task, the robotic
hand moves to the grasping point and executes the grasp. The action
space includes the eight joints of the Schunk robotic hand, while
the observation space comprises the relative pose, distance, and contact
forces between the hand and the object. A binary reward function assigns
1 for successful grasps and 0 otherwise. Through continuous optimization,
the model improves grasp robustness, ensuring a more stable and reliable
interaction with diverse reconstructed objects.

\section{Experiment\thispagestyle{empty}}

We evaluated the proposed framework on a dual-arm robotic platform,
measuring the grasping success rate to assess its effectiveness. The
volumetric reconstruction performance was evaluated using computer-aided
design (CAD) models of objects from TraceParts\footnote{https://www.traceparts.com/en},
which contains comprehensive objects with precisely details. 

\subsection{Performance of 3D Volumetric Reconstruction }

In order to appraise the R-GAN model, we performed three experiments:
evaluation on AffordPose, evaluation on TraceParts, and comparison
study with two other 3D volumetric reconstruction models.

\begin{figure}[H]
\begin{centering}
\includegraphics[width=1\columnwidth]{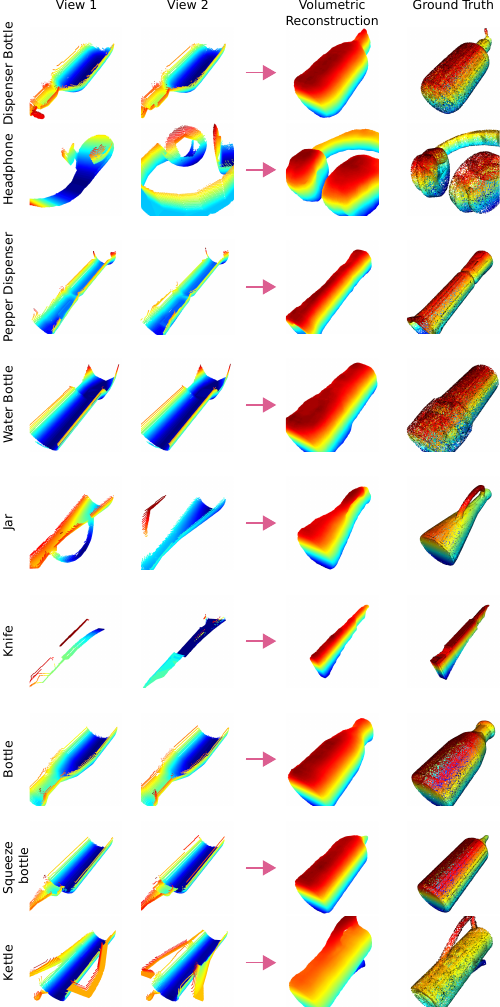}
\par\end{centering}
\caption{Volumetric reconstruction results of AffordPose dataset objects. \label{fig:Sample_Experiment_Results}}
\end{figure}

\begin{figure}[H]
\begin{centering}
\includegraphics[width=1\columnwidth]{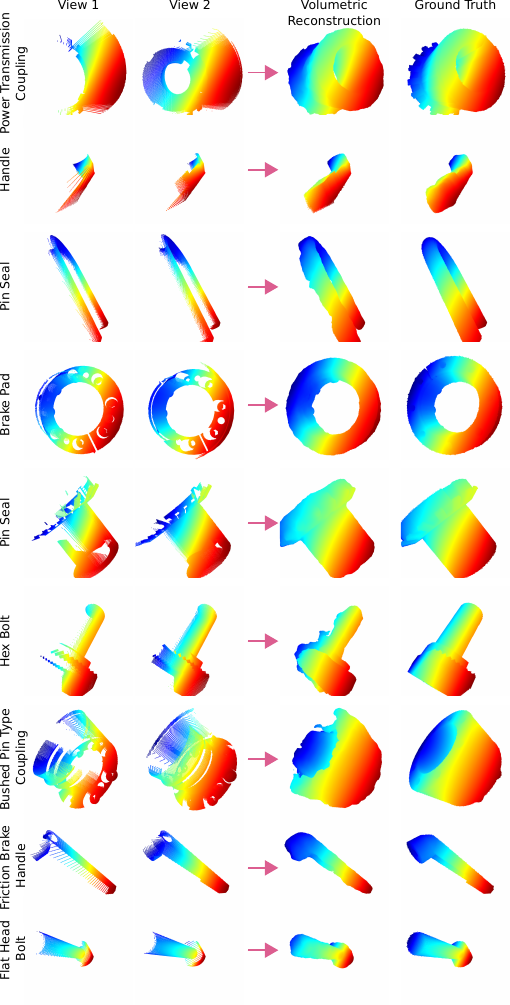}
\par\end{centering}
\caption{Volumetric reconstruction results of objects in the standard dataset.\label{fig:Dataset_Data_Experiment_Results}}
\end{figure}

Volumetric reconstruction results of objects in the AffordPose database
are shown in Fig.~\ref{fig:Sample_Experiment_Results}, and volumetric
reconstruction results of objects in the dataset are shown in Fig.~\ref{fig:Dataset_Data_Experiment_Results}.
Based on the results in Figs.~\ref{fig:Sample_Experiment_Results}
and \ref{fig:Dataset_Data_Experiment_Results}, the proposed model
can successfully reconstruct object volume based on the partial scans.
The proposed model can reconstruct precise structures of objects such
as nuzzle, curly handle, local hollow parts, and convex parts. However,
minor details of parts of the object, which are not included in-depth
scans, are hard to reconstruct, as the discriminator neglected relatively
smaller parts of an object. The absence of these small parts in reconstruction
is a common layback of deep learning-based volumetric reconstruction
methods. 

We computed and compared the mean of IoU, Hit-rate (HR), and Accuracy
of each category shown in the Table~\ref{tab:Projection_Difference-2}.
The IoU was calculated between the 3D fulfilling models that were
generated based on depth scans and CAD models. IoU is computed as
$\mathrm{IoU}=\sum_{i=0}^{N}y_{i}\cap\hat{y_{i}}/\sum_{i=0}^{N}y_{i}\cup\hat{y_{i}},$
where $y_{\mathrm{i}}$ was the $i_{\mathrm{th}}$ voxel in the 3D
fulfilling model, and the $\hat{y}_{\mathrm{i}}$ was the $i_{\mathrm{th}}$
voxel in the 3D CAD model. The hit-rate is the rate of the reconstructed
models of the object that, in the region of the ground truths point
clouds. It indicated how complete the reconstruction model is. The
hit-rate is computed as $\mathrm{HR}=1-(\sum_{i=0}^{N}(y_{i}\cap\hat{y_{i}})\oplus\hat{y_{i}}/\sum_{i=0}^{N}y_{i}\cup\hat{y_{i}})$.
The accuracy showed how well the composed model is. It is computed
as $\mathrm{Accuracy}=1-(\sum_{i=0}^{N}(y_{i}\cap\hat{y_{i}})\oplus y_{i}/\sum_{i=0}^{N}y_{i}\cup\hat{y_{i}})$.

There were 125 depth scans were acquired from different angles for
each object in each category in the Blender environment. HR can describe
the completeness of the reconstructed parts in the 3D reconstruction
models. In detail, the higher the HR results are, the more complete
the volumetric reconstruction models are. Accuracy can describe the
correctness of the 3D-composed results. The higher the accuracy is,
the more points are correctly composed. The IoU, HR, and Accuracy
for ``Power transmission'', ``Sealing'', and ``Handles'' of
3DR2N2 and single-view 3D reconstruction \cite{9635939} methods are
estimated, and they are marked with an asterisk $(^{*})$. The IoU,
HR, and Accuracy of the proposed R-GAN model achieved 77.71\%, 80.08\%,
and 97.45\% on average. Based on the results shown in Table \ref{tab:Projection_Difference-2},
the proposed method outperformed other methods in IoU, HR, and Accuracy.
Therefore, the results proved that for reconstructing 3D models, additional
depth scans could achieve better reconstruction results.\thispagestyle{empty}
\begin{table*}[t]
\centering{}\caption{COMPARISON RESULTS IN IOU, HR, AND ACCURACY. \label{tab:Projection_Difference-2}}
\begin{tabular}{l|>{\raggedright}p{1.1cm}>{\raggedright}p{0.8cm}>{\raggedright}p{1.5cm}|>{\raggedright}p{1.1cm}>{\raggedright}p{0.8cm}>{\raggedright}p{1.5cm}|>{\raggedright}p{1.1cm}>{\raggedright}p{0.8cm}>{\raggedright}p{1.5cm}}
\hline 
\textbf{Category}  & \multicolumn{3}{c|}{\textbf{IoU}} & \multicolumn{3}{c|}{\textbf{Hit-Rate}} & \multicolumn{3}{c}{\textbf{Accuracy}}\tabularnewline
\hline 
 & This Paper  & 3DR2N2\ \textbf{ }\cite{choy20163d}  & Single-View\ \cite{9635939}  & This Paper\  & 3DR2N2\ \cite{choy20163d}  & Single-View\ \cite{9635939}  & This Paper  & 3DR2N2\ \cite{choy20163d}  & Single-View\ \cite{9635939}\tabularnewline
\hline 
\multicolumn{1}{>{\raggedright}b{2.2cm}|}{Brake/clutch/coupling} & \textbf{$\boldsymbol{79.9\%}$}  & $64.2\%$  & $70.48\%$  & \textbf{$\boldsymbol{81.01\%}$}  & $65.09\%$  & $71.46\%$  & \textbf{$\boldsymbol{98.90\%}$}  & $64.2\%$  & $70.48\%$\tabularnewline
Linear and rotary motion  & \textbf{$\boldsymbol{81.87\%}$}  & $73.85\%$  & $76.46\%$  & \textbf{$\boldsymbol{88.77\%}$}  & $73.85\%$  & $76.46\%$  & $\boldsymbol{93.11\%}$  & $73.85\%$  & $76.46\%$\tabularnewline
Power transmission  & \textbf{$\boldsymbol{76.31\%}$}  & $62.45\%^{*}$  & $69.25\%^{*}$  & \textbf{$\boldsymbol{76.88\%}$}  & $69.53\%^{*}$  & $67.75\%^{*}$  & \textbf{$\boldsymbol{99.65\%}$}  & $66.25\%^{*}$  & $74.02\%^{*}$\tabularnewline
Sealing  & \textbf{$\boldsymbol{83.77\%}$}  & $68.56\%^{*}$  & $77.13\%^{*}$  & $\boldsymbol{82.68\%}$  & $60.22\%^{*}$  & $72.86\%^{*}$  & \textbf{$\boldsymbol{99.88\%}$}  & $66.40\%^{*}$  & $74.19\%^{*}$\tabularnewline
Handles  & \textbf{$\boldsymbol{73.37\%}$}  & $60.05\%^{*}$  & $67.06\%^{*}$  & \textbf{$\boldsymbol{76.76\%}$}  & $64.76\%^{*}$  & $67.64\%^{*}$  & \textbf{$\boldsymbol{96.63\%}$}  & $64.24\%^{*}$  & $71.78\%^{*}$\tabularnewline
Fastener  & \textbf{$\boldsymbol{71.06\%}$}  & $53.3\%$  & $67.12\%$  & \textbf{$\boldsymbol{74.56\%}$}  & $53.3\%$  & $67.12\%$  & \textbf{$\boldsymbol{96.51\%}$}  & $53.3\%$  & $67.12\%$\tabularnewline
\hline 
Average  & \textbf{$\boldsymbol{77.71\%}$}  & $57.1\%$  & $71.32\%$  & $\boldsymbol{80.08\%}$  & $64.45\%$  & $70.55\%$  & \textbf{$\boldsymbol{97.45\%}$}  & $64.71\%$  & $72.33\%$\tabularnewline
\hline 
\end{tabular}
\end{table*}

Based on the results shown in Table~\ref{tab:Projection_Difference-2},
volumetric reconstruction models in the ``Sealing'' category achieved
the highest IoU, and volumetric reconstruction models in the ``Fastener''
category achieved the lowest IoU. The reason is that the structures
of objects in the ``Sealing'' are relatively simple, are majorly
ring-shaped, and without complex engraved lines. Objects in the ``Fastener''
category are not in the training dataset or the testing dataset, and
their structures are relatively complex, like fillet conjunctions
between rods and heads. Objects in the ``Linear and rotary motion''
and the ``Power transmission'' categories are both cylinder shapes.
However, the accuracy for reconstructed 3D models in the ``Linear
and rotary motion'' is 93.11\%, and it is 6.54\% lower than the accuracy
for reconstructed 3D models in the ``Power transmission'' category.
Objects in the ``Linear and rotary motion'' category are normally
longer than objects in the ``Power transmission'' category. Therefore,
the depth scans have larger missing parts than objects in the ``Power
transmission'' category.

\subsection{Task-Oriented Grasping}

To evaluate the performance of the proposed framework in task-oriented
grasping, four objects were reconstructed and compared with 3D models
from the revised AffordPose knowledge base. The grasping strategy
of the most volume-similar object was applied. To improve efficiency,
the AffordPose dataset was reorganized into four task categories:
handle grasp, wrap grasp, lift, and press. The framework first identified
the relevant task category, then computed volume similarity using
the Chamfer Distance (CD). The object with the smallest CD was selected
for grasping strategy transfer. 
\begin{figure}[H]
\begin{centering}
\includegraphics[width=0.9\columnwidth]{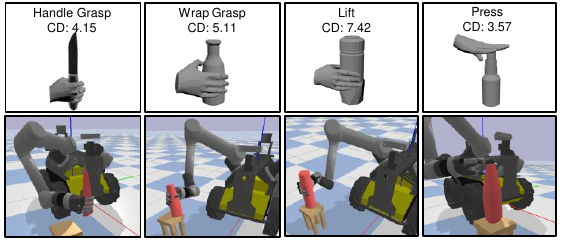}
\par\end{centering}
\caption{Task-oriented grasping. The first row displays the most volume-similar
object from the knowledge base along with its corresponding grasp
strategy. The second row shows the grasping results for the reconstructed
object. \label{fig: graspingexp}}
\end{figure}

Figure \ref{fig: graspingexp} presents the comparison and grasping
results. Experiments showed that, within the same task, higher volume
similarity generally led to a more similar grasping strategy. However,
for different tasks, functional requirements played a more significant
role in determining the appropriate grasp.

A comprehensive reinforcement learning model was train to adapt the
retrieved grasping strategy to the reconstructed object and ensure
stability. For each grasping task, the model selected a grasping point
within a predefined region centered on the retrieved grasping point
(\textpm 3 cm tolerance). The robotic hand move to the grasping point
and executed the grasp. The action space included the eight joints
of the Schunk robotic hand, while the observation space comprised
relative pose, distance, and contact forces between the hand and the
object. A binary reward function assigned 1 for successful grasps
and 0 otherwise. 
\begin{figure}[H]
\begin{centering}
\includegraphics[width=0.8\columnwidth]{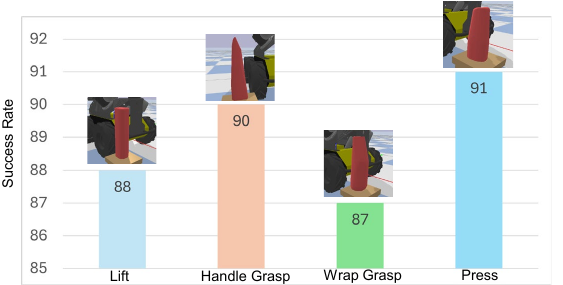}
\par\end{centering}
\caption{Evaluation results for the comprehensive reinforcement learning model.
Each grasping task was performed 100 times, and the success rate was
recorded. \label{fig: graspdata}}
\end{figure}

The model was trained on the four objects over 1,000 episodes and
evaluated in 100 trials, achieving an 89\% testing success rate (Fig.
\ref{fig: graspdata}). Analysis revealed that handle grasping and
pressing had higher success rates. Pressing required only actuator
movement, while handle grasping benefited from high volume similarity
with the knowledge base, ensuring reliable grasping.

Conversely, lifting had lower success rates due to the reconstructed
object's larger size, resulting in a higher CD and reduced grasp stability.
Wrap grasping also performed poorly due to missing fine details in
reconstruction, such as the bottleneck, leading to slipping during
execution.

\section{Conclusion}

This paper presents a task-oriented grasping  framework that integrates
both volumetric and affordance knowledge. The proposed R-GAN model
reconstructs 3D object volumes from limited depth scans, leveraging
LSTM units for sequential learning. By retrieving grasping strategies
from the AffordPose dataset and refining them with PPO, the framework
adapts to unseen objects. Experimental results demonstrate superior
reconstruction accuracy and grasping performance compared to existing
methods. The success rate achieved an average of 89\% over 400 grasping
tasks.

\bibliographystyle{IEEEtran}

\begin{thebibliography}{10}
\providecommand{\url}[1]{#1}
\csname url@samestyle\endcsname
\providecommand{\newblock}{\relax}
\providecommand{\bibinfo}[2]{#2}
\providecommand{\BIBentrySTDinterwordspacing}{\spaceskip=0pt\relax}
\providecommand{\BIBentryALTinterwordstretchfactor}{4}
\providecommand{\BIBentryALTinterwordspacing}{\spaceskip=\fontdimen2\font plus
\BIBentryALTinterwordstretchfactor\fontdimen3\font minus
  \fontdimen4\font\relax}
\providecommand{\BIBforeignlanguage}[2]{{%
\expandafter\ifx\csname l@#1\endcsname\relax
\typeout{** WARNING: IEEEtran.bst: No hyphenation pattern has been}%
\typeout{** loaded for the language `#1'. Using the pattern for}%
\typeout{** the default language instead.}%
\else
\language=\csname l@#1\endcsname
\fi
#2}}
\providecommand{\BIBdecl}{\relax}
\BIBdecl

\bibitem{rodriguez2012caging}
A.~Rodriguez, M.~T. Mason, and S.~Ferry, ``From caging to grasping,'' \emph{The
  International Journal of Robotics Research}, vol.~31, no.~7, pp. 886--900,
  2012.

\bibitem{weisz2012pose}
J.~Weisz and P.~K. Allen, ``Pose error robust grasping from contact wrench
  space metrics,'' in \emph{2012 IEEE international conference on robotics and
  automation}.\hskip 1em plus 0.5em minus 0.4em\relax IEEE, 2012, pp. 557--562.

\bibitem{goldfeder2011data}
C.~Goldfeder and P.~K. Allen, ``Data-driven grasping,'' \emph{Autonomous
  Robots}, vol.~31, no.~1, pp. 1--20, 2011.

\bibitem{mahler2017dex}
J.~Mahler, J.~Liang, S.~Niyaz, M.~Laskey, R.~Doan, X.~Liu, J.~A. Ojea, and
  K.~Goldberg, ``Dex-net 2.0: Deep learning to plan robust grasps with
  synthetic point clouds and analytic grasp metrics,'' \emph{arXiv preprint
  arXiv:1703.09312}, 2017.

\bibitem{varley2017shape}
J.~Varley, C.~DeChant, A.~Richardson, J.~Ruales, and P.~Allen, ``Shape
  completion enabled robotic grasping,'' in \emph{2017 IEEE/RSJ international
  conference on intelligent robots and systems (IROS)}.\hskip 1em plus 0.5em
  minus 0.4em\relax IEEE, 2017, pp. 2442--2447.

\bibitem{runz2020frodo}
M.~Runz, K.~Li, M.~Tang, L.~Ma, C.~Kong, T.~Schmidt, I.~Reid, L.~Agapito,
  J.~Straub, S.~Lovegrove \emph{et~al.}, ``Frodo: From detections to 3d
  objects,'' in \emph{Proceedings of the IEEE/CVF Conference on Computer Vision
  and Pattern Recognition}, 2020, pp. 14\,720--14\,729.

\bibitem{9562073}
H.~Li, Y.~Zhang, Y.~Li, and H.~He, ``Learning task-oriented dexterous grasping
  from human knowledge,'' in \emph{2021 IEEE International Conference on
  Robotics and Automation (ICRA)}.\hskip 1em plus 0.5em minus 0.4em\relax IEEE,
  2021, pp. 6192--6198.

\bibitem{yin2021modeling}
H.~Yin, A.~Varava, and D.~Kragic, ``Modeling, learning, perception, and control
  methods for deformable object manipulation,'' \emph{Science Robotics},
  vol.~6, no.~54, p. eabd8803, 2021.

\bibitem{9635939}
F.~Yan, D.~Wang, and H.~He, ``Comprehension of spatial constraints by neural
  logic learning from a single rgb-d scan,'' in \emph{2021 IEEE/RSJ
  International Conference on Intelligent Robots and Systems (IROS)}, 2021, pp.
  9008--9013.

\bibitem{green2008human}
S.~A. Green, M.~Billinghurst, X.~Chen, and J.~G. Chase, ``Human-robot
  collaboration: A literature review and augmented reality approach in
  design,'' \emph{International journal of advanced robotic systems}, vol.~5,
  no.~1, p.~1, 2008.

\bibitem{yi2014survey}
G.~Yi, L.~Jianxin, Q.~Hangping, and W.~Bo, ``Survey of structure from motion,''
  in \emph{Proceedings of 2014 International Conference on Cloud Computing and
  Internet of Things}.\hskip 1em plus 0.5em minus 0.4em\relax IEEE, 2014, pp.
  72--76.

\bibitem{silveira2008efficient}
G.~Silveira, E.~Malis, and P.~Rives, ``An efficient direct approach to visual
  slam,'' \emph{IEEE transactions on robotics}, vol.~24, no.~5, pp. 969--979,
  2008.

\bibitem{10.1145/2508363.2508374}
\BIBentryALTinterwordspacing
M.~Nie\ss{}ner, M.~Zollh\"{o}fer, S.~Izadi, and M.~Stamminger, ``Real-time 3d
  reconstruction at scale using voxel hashing,'' \emph{ACM Trans. Graph.},
  vol.~32, no.~6, nov 2013. [Online]. Available:
  \url{https://doi.org/10.1145/2508363.2508374}
\BIBentrySTDinterwordspacing

\bibitem{sorkine2004laplacian}
O.~Sorkine, D.~Cohen-Or, Y.~Lipman, M.~Alexa, C.~R{\"o}ssl, and H.-P. Seidel,
  ``Laplacian surface editing,'' in \emph{Proceedings of the 2004
  Eurographics/ACM SIGGRAPH symposium on Geometry processing}, 2004, pp.
  175--184.

\bibitem{kazhdan2006poisson}
M.~Kazhdan, M.~Bolitho, and H.~Hoppe, ``Poisson surface reconstruction,'' in
  \emph{Proceedings of the fourth Eurographics symposium on Geometry
  processing}, vol.~7, 2006.

\bibitem{Jian_2023_ICCV}
J.~Jian, X.~Liu, M.~Li, R.~Hu, and J.~Liu, ``Affordpose: A large-scale dataset
  of hand-object interactions with affordance-driven hand pose,'' in
  \emph{Proceedings of the IEEE/CVF International Conference on Computer Vision
  (ICCV)}, October 2023, pp. 14\,713--14\,724.

\bibitem{wang2014robust}
K.~Wang, G.~Zhang, and H.~Bao, ``Robust 3d reconstruction with an rgb-d
  camera,'' \emph{IEEE Transactions on Image Processing}, vol.~23, no.~11, pp.
  4893--4906, 2014.

\bibitem{wang2015mobile}
W.~Wang, K.~Yamakawa, K.~Hiroi, K.~Kaji, and N.~Kawaguchi, ``A mobile system
  for 3d indoor mapping using lidar and panoramic camera,'' in \emph{Adjunct
  Proceedings of the 2015 ACM International Joint Conference on Pervasive and
  Ubiquitous Computing and Proceedings of the 2015 ACM International Symposium
  on Wearable Computers}, 2015, pp. 337--340.

\bibitem{choy20163d}
C.~B. Choy, D.~Xu, J.~Gwak, K.~Chen, and S.~Savarese, ``3d-r2n2: A unified
  approach for single and multi-view 3d object reconstruction,'' in
  \emph{European conference on computer vision}.\hskip 1em plus 0.5em minus
  0.4em\relax Springer, 2016, pp. 628--644.

\end{thebibliography}

\end{document}